\documentclass[conference]{IEEEtran}
\IEEEoverridecommandlockouts
\usepackage{cite}
\usepackage{amsmath,amssymb,amsfonts}
\usepackage{algorithmic}
\usepackage{graphicx}
\usepackage{textcomp}
\usepackage{xcolor}
\usepackage{multirow}
\usepackage{booktabs}
\usepackage{threeparttable}
\usepackage{tabularx}
\newsavebox{\tablebox}
\usepackage{algorithm}
\DeclareMathOperator*{\argmin}{arg\,min}
\usepackage{array}
\newcolumntype{L}[1]{>{\raggedright\arraybackslash}p{#1}}

\def\BibTeX{{\rm B\kern-.05em{\sc i\kern-.025em b}\kern-.08em
    T\kern-.1667em\lower.7ex\hbox{E}\kern-.125emX}}
\begin{document}

\title{Unifying Unsupervised Domain Adaptation and Zero-Shot Visual Recognition
}

\author{\IEEEauthorblockN{ Qian Wang\textsuperscript{1}, Penghui Bu\textsuperscript{1,2} and Toby P. Breckon\textsuperscript{1}}
\IEEEauthorblockA{\textsuperscript{1}\textit{Department of Computer Science, Durham University, UK}\\
\textsuperscript{2}\textit{ School of Mechanical Engineering, Xi'an Jiaotong University, China} }}

\maketitle

\begin{abstract}
Unsupervised domain adaptation aims to transfer knowledge from a source domain to a target domain so that the target domain data can be recognized without any explicit labelling information for this domain. One limitation of the problem setting is that testing data, despite having no labels, from the target domain is needed during training, which prevents the trained model being directly applied to classify unseen test instances. We formulate a new cross-domain classification problem arising from real-world scenarios where labelled data is available for a subset of classes (known classes) in the target domain, and we expect to recognize new samples belonging to any class (known and unseen classes) once the model is learned. This is a generalized zero-shot learning problem where the side information comes from the source domain in the form of labelled samples instead of class-level semantic representations commonly used in traditional zero-shot learning. We present a unified domain adaptation framework for both unsupervised and zero-shot learning conditions. Our approach learns a joint subspace from source and target domains so that the projections of both data in the subspace can be domain invariant and easily separable. We use the supervised locality preserving projection (SLPP) as the enabling technique and conduct experiments under both unsupervised and zero-shot learning conditions, achieving state-of-the-art results on three domain adaptation benchmark datasets: Office-Caltech, Office31 and Office-Home.
\end{abstract}


\section{Introduction}
Training a visual recognition model for image classification requires large amount of annotated data which hinders its application in many real-world scenarios where  very few or no labelled images exist in the target domain. One solution is to use the labelled data from a different domain for training and apply the model to the target task. For example, if we assume our task is to classify artistic images for which we do not have much labelled data, it maybe conversely easier to get access to many labelled natural images. Training a classifier with the natural images and applying it directly to the artistic data suffers due to the inherent domain shift. To address this problem, domain adaptation approaches have been proposed to transform original features so that the transformed source and target features can be aligned \cite{long2013transfer, long2014transfer, sun2016return, zhang2017joint, ghifary2017scatter, sun2017correlation, wang2018visual}. Recently, deep feature learning approaches have drawn much attention by using end-to-end deep models to learn domain-invariant features from different domains \cite{ganin2015unsupervised, long2015learning, ganin2016domain, long2016unsupervised, long2017deep, chen2018joint, pei2018multi, zhang2018collaborative, long2018conditional}.

Although significant efforts have been made to address the domain adaptation problem, most focus on unsupervised domain adaptation for which the target domain data are assumed to be accessible for learning although no labelling information is available. This is a strong assumption and hinders a direct application of the learned model to out-of-sample classification. In many real-world scenarios, it is easier to get some labelled examples for some classes than the others in the target domain. With these limited labelled samples from the target domain, it is worth investigating the possibility of domain adaptation without accessing the testing samples. For this purpose, we formulate a novel domain adaptation problem under the zero-shot learning (ZSL) \cite{wang2017zero} condition and subsequently propose a viable solution to it. 

Specifically, we present a unified framework for visual domain adaptation under both unsupervised and zero-shot learning conditions. Our approach aims to learn a subspace in which the domain and target data can be aligned and well-separated. To this end, a supervised locality preserving projection (LPP) \cite{he2004locality, wang2017zero} is employed as an enabling technique for subspace learning. For unsupervised domain adaptation, we propose a confidence-aware pseudo label selection scheme to gradually align the domains in an iterative learning strategy. To evaluate the effectiveness of the proposed approaches, we conduct experiments on commonly used datasets for domain adaptation, achieving state-of-the-art performance.

\section{Problem Formulation}\label{sec:prob}
To facilitate our presentation in the following sections, we firstly formulate domain adaptation problems under the unsupervised learning and zero-shot learning conditions respectively. Given a labelled dataset $\mathcal{D}^s = \{(x^s_i,y^s_i)\}, i = 1,2,...,n^s$ from the source domain $\mathcal{S}$, $x^s_i \in \mathbb{R}^{d^s}$ represents the feature vector of $i$-th training example in the source domain, $d^s$ is the feature dimension in the source domain and $y^s_i \in \mathcal{Y}^s$ denotes the corresponding label. For the unsupervised domain adaptation problem, the task is to classify an unlabelled dataset $\mathcal{D}^t = \{x^t_i\}, i=1,2,...,n^t$ from the target domain $\mathcal{T}$, where $x^t_i \in \mathbb{R}^{d^t}$ represents the feature vector in the target domain and $d^t$ is the dimensionality of features. The target label space $\mathcal{Y}^t$ is equal to the source label space $\mathcal{Y}^s$. It is assumed that both the labelled source domain data $\mathcal{D}^s$ and the unlabelled target domain data $\mathcal{D}^t$ are available for unsupervised domain adaptation learning.

For a zero-shot learning condition, we have $\mathcal{D}^s$ as above as well as a labelled dataset $\mathcal{D}^{tl}=\{(x_i^{tl},y_i^{tl})\},i=1,2,...,n^{tl}$ from the target domain $\mathcal{T}$. $x_i^{tl} \in \mathbb{R}^{d^t}$ and $y_i^{tl} \in \mathcal{Y}^{tl}$ are the feature and label of the $i$-th labelled example respectively. The task is to classify any given new instance $x^{t}$ from the target domain by learning an inference model $y=f(x^t) \in \mathcal{Y}^t$. It is noteworthy that $\mathcal{Y}^{tl} \subset \mathcal{Y}^t = \mathcal{Y}^s$, that is, only a subset of the target labels have labelled training examples available during learning, while the instance to classify could belong to any class in the whole target label space.

\begin{figure}
	\centering
	{\includegraphics[width = 3.5in]{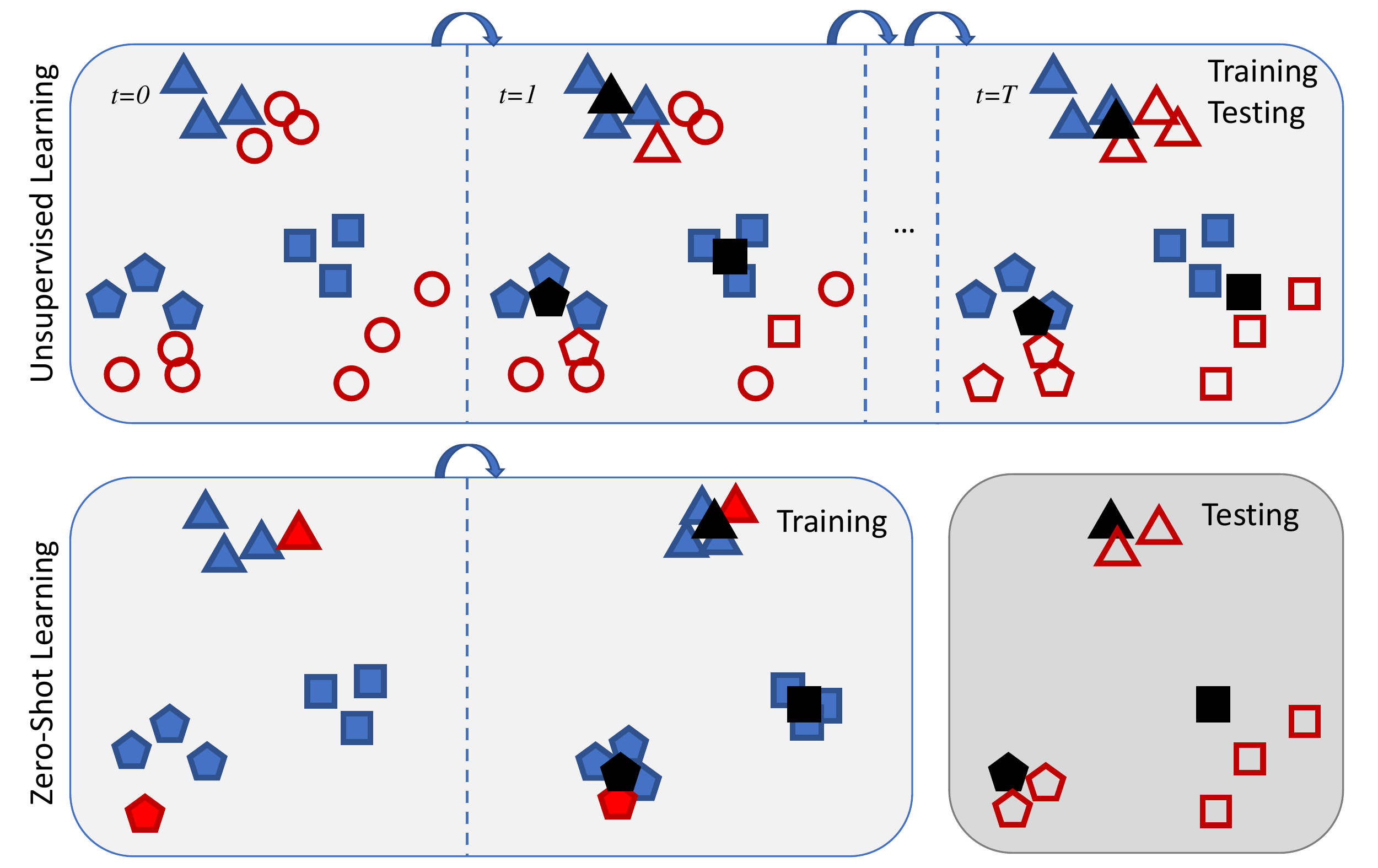}}
	{\caption{The proposed framework of domain adaptation for unsupervised (upper) and zero-shot learning (bottom).
			\newline
			{\color{blue}Blue} and {\color{red}Red} markers represent data from source and target domains respectively. The Black markers represent learned class-level representations. The shapes of ``triangle", ``diamond" and ``square" denote three different classes whilst the shape of ``circle" represent unlabelled samples. Filled and hollow markers represent ground truth labelling and predictions respectively. The main difference between unsupervised and zero-shot learning conditions is the access of target data for training as denoted by the presence and absence of the hollow markers in the left of the two conditions illustrated above.
			}
		\label{figure_framework}}
\end{figure}

The domain adaptation problem under zero-shot learning conditions is relatively new and under-explored. It differs from the traditional unsupervised domain adaptation in two ways. On one hand, unsupervised domain adaptation assumes there is no labelled data from target domain, while in the zero-shot learning problem it is assumed there exist some labelled examples from the target domain although the labelled examples are only restricted to a subset of the whole target label space. On the other hand, the testing samples in the zero-shot learning condition are not available during training while in the unsupervised condition they are used for training together with other labelled data. An illustration for the two conditions is shown in Figure \ref{figure_framework}.
In addition, the zero-shot learning condition is also different from supervised domain adaptation \cite{tzeng2015simultaneous,motiian2017unified} and semi-supervised domain adaptation \cite{xiao2015feature,ding2018semi} where labelled examples in the target domain are assumed to be available for all classes (i.e. $\mathcal{Y}^{tl} = \mathcal{Y}^t$).

Domain adaptation under zero-shot learning condition poses different challenge from other domain adaptation problems. It could be easy to classify target data from known classes but quite difficult for those from unseen classes since models learned under this condition could bias to known classes and mistakenly classify all target data as known classes.

\section{Related Work}\label{sec:related}
In this section, we firstly review existing work related to unsupervised domain adaptation. Subsequently, we briefly describe zero-shot learning problems, how we formulate the domain adaptation problem under the zero-shot learning condition and existing work related to ours.
\subsection{Unsupervised Domain Adaptation}
Unsupervised domain adaptation has attracted much attention in recent years. Existing approaches to unsupervised domain adaptation in literature can be roughly categorized into two groups: \textit{feature transformation approaches} \cite{long2013transfer, long2014transfer, sun2016return, zhang2017joint, ghifary2017scatter, sun2017correlation, wang2018visual} and \textit{deep feature learning approaches} \cite{ganin2015unsupervised, long2015learning, ganin2016domain, long2016unsupervised, long2017deep, chen2018joint, pei2018multi, zhang2018collaborative, long2018conditional}. 

Feature transformation approaches aim to transform the source domain and/or target domain features such that transformed source and target domain data can be aligned. As such the classifier learned from labelled source data can be directly applied to target data. Usually linear transformations are used by learning the projection matrices with different optimization objectives and a kernel trick can help to explore the non-linear relations between source and target domain data if necessary. The most commonly employed objective for unsupervised domain adaptation is to align data distributions in source and target domains \cite{long2013transfer, long2014transfer}. For this purpose, Maximum Mean Discrepancy (MMD) based distribution matching has been used to reduce differences of the marginal distributions \cite{long2014transfer}, conditional distributions or both \cite{long2013transfer, wang2018visual}. 
Manifold Embedded Distribution Alignment (MEDA) \cite{wang2018visual} learns a domain-invariant classifier based on the transformed features where the transformation aims to align both the marginal and conditional distributions with quantitative account for their relative importance. 
Joint Geometrical and Statistical Alignment (JGSA) \cite{zhang2017joint} learns two coupled projections that project the source and target domain data into a joint subspace where the geometrical and distribution shifts are reduced simultaneously. 
Apart from the distribution alignment, recent feature transformation based approaches also promote the discriminative properties in the transformed features. Scatter Component Analysis (SCA) \cite{ghifary2017scatter} aims to learn a feature transformation such that the transformed data from different domains have similar scattering and the labelled data are well separated. A Linear Discriminant Analysis (LDA) framework was proposed in \cite{lu2018embarrassingly} by learning class-specific projections. Similarly, Li et al. \cite{li2018domain} proposed an approach to feature transformation towards Domain Invariant and Class Discriminative (DICD) features.

The proposed approach in this paper falls under this category since it learns a joint subspace from source and target domains. To learn the projection matrix transforming both source and target features into the joint subspace, we take advantage of pseudo labels of target data and iteratively update the projection matrix with the combination of labelled source data and pseudo-labelled target data. This strategy of using the pseudo labels of the target domain data with iterative learning has been employed in many approaches \cite{long2013transfer, long2014transfer, zhang2017joint, wang2018visual}. In contrast to the iterative learning in the existing approaches which use all the pseudo-labelled target data, we select a part of target data which have been labelled with relatively higher confidence whilst ignore the ones with lower confidence in each iteration (see Section \ref{sec:method_uda}).

Deep feature learning approaches to domain adaptation were inspired by the success of deep Convolutional Neural Networks (CNN) in visual recognition \cite{lecun2015deep}. Attempts have been made to take advantage of the powerful representation learning capability of CNN combined with a variety of feature learning objectives. Most deep feature learning approaches aim to learn domain-invariant features from raw image data in source and target domains in an end-to-end framework. Specifically, the objectives of feature transformation approaches have been incorporated in the deep learning models. To learn the domain-invariant features through a deep CNN, the gradient reversal layer was proposed in \cite{ganin2015unsupervised} and used in other deep feature learning approaches \cite{ganin2016domain, pei2018multi, zhang2018collaborative} as well. The gradient reversal layer connects the feature extraction layers and the domain classifier layers. During backpropagation, the gradients of this layer multiplies a certain negative constant to ensure the feature distributions over two domains are made similar (as indistinguishable as possible for the domain classifier). Deep Adaptation Networks (DAN) \cite{long2015learning} and Residual Transfer Network (RTN) \cite{long2016unsupervised} aim to learn transferable features from two domains by matching the domain distributions of multiple hidden layer features based on MMD. Deep CORAL \cite{sun2016deep} integrates the idea of CORAL \cite{sun2016return} into a deep CNN framework to learn features with favoured properties (i.e. aligned correlations over source and target distributions for multiple layer activations). These approaches only consider the alignment of marginal distributions and cannot ensure the separability of target data. Deep Reconstruction Classification Network (DRCN) \cite{ghifary2016deep} trains a feature learning model using labelled source data and unlabelled target data in the supervised and unsupervised learning manners respectively. More recently, the prevalent Generative Adversarial Network (GAN) loss has been employed in Adversarial Discriminative Domain Adaption (ADDA) \cite{tzeng2017adversarial} with promising results. 

Though deep learning based approaches are able to train the models in an end-to-end way, their performance on benchmark datasets has not outperformed the feature transformation based approaches especially when the deep features are used for feature transformation itself.

\subsection{Zero-Shot Learning} \label{sec:zsl}
Zero-shot learning (ZSL) aims to recognize novel classes by transferring knowledge learned from known classes to unseen classes \cite{wang2017zero}. ZSL has attracted much attention since it provides a promising solution to the sparse labelling issues in real world applications. In traditional zero-shot visual recognition tasks, the source domain data are usually of a different modality such as human-defined class attributes and large corpus hence it suffers from the semantic gap between visual and semantic representations \cite{wang2017zero}. Since the domain adaptation problem under the zero-shot learning condition formulated in Section \ref{sec:prob} assumes both source and target data are from visual domain, the semantic gap issue suffered in traditional zero-shot learning tasks can be alleviated though the domain shift still exists. Traditional ZSL approaches can only tackle the class-level semantic representations (e.g., attributes and word vectors) even the source domain data come with multiple labelled examples \cite{wang2017alternative}. As a result, most existing ZSL methods are not ready to be directly applied in our proposed problem.

Domain adaptation under the zero-shot learning condition has been investigated in \cite{saenko2010adapting} and \cite{tzeng2015simultaneous}. However, this work only focused on the conventional zero-shot learning \cite{xian2018zero} where the test instances are restricted to be only from unseen classes. Our work aims to address the generalized zero-shot learning problem \cite{xian2018zero} which arises from a more realistic situation where test instances can belong to any class (i.e. either known or unseen classes).

\section{Method}\label{sec:method}
The proposed method aims to learn a subspace from the domain and target features so that the transformed features in the subspace are domain-invariant and well-separated. We have many options for this purpose including linear discriminant analysis (LDA) and supervised locality preserving projection (SLPP). According to \cite{wang2017zero}, SLPP has favourable properties that data structures can be preserved after projection hence avoiding overfitting to the training data. It is therefore appropriate for the problems where test data and training data have different distributions. For unsupervised domain adaptation, there is no labelling information available for target domain data. To address this issue, we use pseudo labels generated by a classifier (e.g. nearest neighbour). To avoid the wrongly labelled target instances undermining the subspace learning process by propagating the errors to the next iteration, we propose a Confidence-Aware Pseudo Label Selection (CAPLS) scheme. For the zero-shot learning condition, we use the labelled source data and labelled target data to learn the subspace and project both source and target data into the subspace in which the out-of-sample classification can be done using nearest neighbour (to the learned class-level representations). In the following subsections, we describe each component in detail.
\subsection{Data Preprocessing}\label{sec:method_pre}
As the first step of our approach, $l$2 normalization is applied to all data as follows:
\begin{equation}\label{eq:pre}
\hat{x} = x/||x||_2.
\end{equation}
The use of $l$2 normalization results in the data points distributed on the surface of a hyper-sphere which will help to align data from different domains \cite{wang2017zero}. Our experimental results in this study also provide empirical evidence that sample normalization is beneficial to superior performance. Besides, feature normalization (e.g., Z-score normalization) might be needed depending on the features used, which, however, is not a must for deep features employed in our experiments.
\subsection{Joint Subspace Learning}\label{sec:method_jsl}
We aim to learn a subspace to which the source and target domain data can be projected by a projection matrix and the projected data are domain-invariant and well-separated. To this end, the supervised locality preserving projection \cite{wang2017zero} is employed.
We denote the projection matrix as $P\in \mathbb{R}^{d^s\times d}$, where $d$ is the dimensionality of the subspace and $d^s$ is assumed to be equal to $d^t$ in the following presentation for convenience without loss of generality. 

In SLPP, the following cost function is employed to learn the projection matrix $P$:
\begin{equation}
\label{eq:cost}
L(P;W,X^l) = \sum_{i,j} || P^T x_i - P^T x_j||_2^2 W_{ij}
\end{equation}
where $x_i$ is the $i$-th column of the labelled data matrix $X^l$ which is a collection of labelled data in both source and target domains and each column vector represents an instance. The similarity matrix $W$ is defined as follows:
\begin{equation}
\label{eq:sim}
W_{ij} = \left \{
\begin{array}{ll}
1, & y_i = y_j,\\
0 ,&  otherwise
\end{array}
\right.
\end{equation}
That is, $W_{ij}$ is set to 1 when $x_i$ and $x_j$ have the same label regardless of which domain they are from, otherwise the value is set 0. This is different from the original SLPP in \cite{wang2017zero} where the distances between samples are considered to construct the similarity matrix $W$. Here we ignore the within-class sample distances since we aim to align the source and domain data in the learned subspace. It has been proved empirically that Eq.(\ref{eq:sim}) is sufficient to capture the intrinsic data structures to learn a domain-invariant yet discriminative subspace by minimizing the cost function defined in Eq.(\ref{eq:cost}).

Minimizing the cost function in Eq.(\ref{eq:cost}) enables the instances of the same class stay close to each other in the learned subspace no matter whether they are from the same domain or different domains. Following the treatment in \cite{he2004locality, wang2017zero}, the objective can be rewritten in the following form:
\begin{equation}
\label{eq:obj}
\max_{P} \frac{Tr(P^TX^lD {X^l}^T P)}{Tr(P^T(X^lL {X^l}^T + I)P)}
\end{equation}
where $L=D-W$ is the laplacian matrix, $D$ is a diagonal matrix with $D_{ii}=\sum_j W_{ij}$ and the regularization term $Tr(P^T P)$ is added for penalizing extreme values in the projection matrix $P$.

The problem defined in Eq.(\ref{eq:obj}) is equivalent to the following generalized eigenvalue problem:
\begin{equation}
\label{eq:eig}
X^lD{X^l}^Tp = \lambda (X^lL{X^l}^T+I)p,
\end{equation}
solving the generalized eigenvalue problem gives the optimal solution $P=[p_1, ..., p_d]$ where $p_1,...,p_d$ are the eigenvectors corresponding to the largest $d$ eigenvalues.

\subsection{Recognition in Subspace}\label{sec:method_rec}
Once the joint subspace is learned, we can project data from either source or target domain into the subspace by:
\begin{equation}\label{eq:prj_src}
z_i = P^T x_i
\end{equation}
where $z_i$ is the projection of $x_i$ in the subspace.

To facilitate the separability of data projected into the subspace, we follow \cite{wang2017zero} and apply the centralization (i.e. mean subtraction) and $l$2 normalization to all the projections:
\begin{equation}\label{eq:sub}
z \leftarrow z-\bar{z},
\end{equation} 
\begin{equation}\label{eq:l2n}
z \leftarrow z/||z||_2,
\end{equation}
where $\bar{z}$ is the mean of all the projected training data (i.e., all the source and target data for unsupervised domain adaptation condition; source data and labelled target data for zero-shot learning condition).

Given any instance $x^t$ from the target domain, we now predict its corresponding label $y^t$. We firstly project the instance from target domain into the subspace by Eq.(\ref{eq:prj_src}) and then apply the centralization and normalization by Eq.(\ref{eq:sub}) and Eq.(\ref{eq:l2n}).
The label of a target instance is then predicted with:
\begin{equation}\label{eq:rec}
\hat{y}^t = \argmin_y ||z^t-\bar{z}_y||_2, y\in \mathcal{Y}^t,
\end{equation}
where 
\begin{equation}\label{eq:avg}
\bar{z}_y = \frac{\sum_i z^s_i \delta(y,y_i^s) + \sum_j z_j^{lt} \delta(y,y_j^{lt})}{\sum_i \delta(y,y_i^s) + \sum_j \delta(y,y_j^{lt})}
\end{equation}
is the mean vector of the projected source data whose labels are $y$, $\delta(y,y_i)=1$ if $y=y_i$ and 0 otherwise. Note that the class means $\bar{z}_y$ are calculated using only labelled source data for unsupervised domain adaptation and labelled target data are also used for zero-shot learning problem. Following \cite{wang2017zero}, we apply $l$2 normalization to $\bar{z}_y$ before using them in Eq.(\ref{eq:rec}).

To this point, it is straightforward to apply the proposed approach to the zero-shot learning condition and the algorithm is summarized in Algorithm \ref{alg:zsl}. For unsupervised domain adaptation, however, we only have labelled data from source domain and labelled target data are needed for domain alignment. To this end, we use pseudo-labelled target instances and the iterative learning strategy described in the following subsection.

\begin{algorithm}[tb]
	\caption{Domain Adaptation Under Zero-Shot Learning condition}
	\label{alg:zsl}
	\renewcommand{\algorithmicrequire}{\textbf{Input:}}
	\renewcommand{\algorithmicensure}{\textbf{Output:}}
	\begin{algorithmic}[1]
		\REQUIRE The labelled source data $\mathcal{D}^s = \{(x^s_i,y^s_i)\}, i = 1,2,...,n^s$ and labelled target data $\mathcal{D}^{tl}=\{(x_i^{tl},y_i^{tl})\},i=1,2,...,n^{tl}$, dimensionality of subspace $d$, test instance from target domain $x^t$.
		\ENSURE The projection matrix $P$ and predicted label $\hat{y}^t$.
		\STATE Do data pre-processing using Eq.(\ref{eq:pre});
		\STATE Learn the joint subspace (projection matrix $P$) using Eq.(\ref{eq:eig});
		\STATE Predict the label $\hat{y}^t$ for test instance $x^t$ using Eq.(\ref{eq:rec}).
	\end{algorithmic}
\end{algorithm}

\subsection{Unsupervised Domain Adaptation Using CAPLS}\label{sec:method_uda}
Our domain adaptation framework based on joint subspace learning (c.f. Section \ref{sec:method_jsl}) requires labelled data from both source and target domains for domain alignment. However, for unsupervised domain adaptation problem, we do not have any labelled target data. To address issue, as mentioned above, we use pseudo labelled target for domain-invariant subspace learning. Specifically, we learn a projection matrix $P_0$ using only labelled source data and get the pseudo labels of all the target data using Eq.(\ref{eq:rec}). Once the target data are labelled, we combine the pseudo-labelled target data with the labelled source data and relearn the projection matrix $P$. This process is repeated for multiple times, as a result, the learned subspace becomes more domain-invariant and discriminative until convergence. 

One drawback of the iterative learning strategy used in most exiting approaches is that the classification errors in the early iterations will be propagated to the later iterations thus leading the algorithm to a sub-optimal solution. To alleviate this issue, we propose a confidence-aware pseudo label selection scheme. Instead of using all the pseudo-labelled target data, we select a portion of them with high confidence to combine with labelled source data for the next iteration learning.

Specifically, we transform the distance from a given test instance $z$ to the $i$-th class representation $\bar{z}_i$ (i.e.  $d_i=||z-\bar{z}_i||_2$ in Eq.(\ref{eq:rec}) into probability $q_i$ using the following softmax function:
\begin{equation}\label{eq:sft}
q_i = \frac{e^{-d_i}}{\sum_{i=1}^{|\mathcal{Y}^s|} e^{-d_i}}
\end{equation}
where $|\mathcal{Y}^s|$ is the number of labels of the source data (also that of the target data). $q_i$ denotes the probability that the given test instance belongs to the $i$-th class. 

Now we use $Q \in \mathbb{R}^{n^t \times |\mathcal{Y}^s|}$ to collectively denote the predicted probability matrix of all the target data and $Q_{ij}$ denotes the probability of $i$-th target instance belong to $j$-th class. For each class, there is no doubt that the target instances labelled as this class with higher probabilities are more likely correctly labelled, hence they should be selected to participate in the next iteration of learning. As a result, in $t$-th iteration, we select top $t/T$ percent pseudo-labelled target instances for each class as trustable ones for the next-iteration learning, where $T$ is the total number of iteration. It is noteworthy that the selection is class-wise so that there exists pseudo-labelled target data selected for each class. The complete unsupervised domain adaptation approach using CAPLS is summarized in Algorithm \ref{alg:uda} whose time complexity is $\mathcal{O}(T(d^3+dn^2))$.
\begin{algorithm}[tb]
	\caption{Unsupervised Domain Adaptation Using CAPLS}
	\label{alg:uda}
	\renewcommand{\algorithmicrequire}{\textbf{Input:}}
	\renewcommand{\algorithmicensure}{\textbf{Output:}}
	\begin{algorithmic}[1]
		\REQUIRE The labelled source data $\mathcal{D}^s = \{(x^s_i,y^s_i)\}, i = 1,2,...,n^s$ and unlabelled target data $\mathcal{D}^{t}=\{(x_i^t\},i=1,2,...,n^{t}$, dimensionality of subspace $d$, number of iteration $T$.
		\ENSURE The projection matrix $P$ and predicted label $\{\hat{y}^t\}$.
		\STATE Initialize $t=0$;
		\STATE Do data pre-processing using Eq.(\ref{eq:pre});
		\STATE Learn the projection $P_0$ using only $\mathcal{D}^s$;
		\STATE Assign pseudo labels for all target data using Eq.(\ref{eq:rec});
		\WHILE {$t\leq T$}
		\STATE $t \leftarrow t+1$;
		\STATE Select top $t/T$ percent trustful pseudo-labelled target data for each class;
		\STATE Learn $P_t$ using $\mathcal{D}^s$ and selected pseudo-labelled target data;
		\STATE Update pseudo labels for all target data using Eq.(\ref{eq:rec}).
		\ENDWHILE
	\end{algorithmic}
\end{algorithm}

\section{Experiment}\label{sec:experiment}

\begin{table*}[!htbp]
	\centering
	{
		\centering
		\caption[]{Classification Accuracy (\%) on Office-Caltech dataset for unsupervised domain adaptation. The feature transformation approaches use Decaf6 features. Columns display results of source $\to$ target pairs.\\}
		\label{table:uda_o10}
		\begin{lrbox}{\tablebox}
			\begin{tabular}{cccccccccccccc}
				\hline
				Method & C$\to$A & C$\to$W & C$\to$D & A$\to$C&A$\to$W & A$\to$D&W$\to$C & W$\to$A & W$\to$D & D$\to$C & D$\to$A & D$\to$W & Average \\ \hline
				DCORAL\cite{sun2016deep} & 92.4 & 91.1 & 91.4 & 84.7& - & - & 79.3 & - & - & 82.8 & - & - & -\\
				DDC\cite{tzeng2014deep}  & 91.9 & 85.4 & 88.8 & 85.0 & 86.1 & 89.0 & 78.0 & 84.9 & 100.0 & 81.1 & 89.5 & 98.2 & 88.2\\
				DAN\cite{long2015learning}  & 92.0 & 90.6 & 89.3 & 84.1 & \textbf{91.8} & 91.7 & 81.2 & 92.1 & 100.0 & 80.3 & 90.0 & 98.5 & 90.1\\
				\hline
				CORAL\cite{sun2017correlation}& 92.0 & 80.0 & 84.7 & 83.2 & 74.6 & 84.1 & 75.5 & 81.2 & 100.0 & 76.8 & 85.5 & 99.3 & 84.7\\
				SCA\cite{ghifary2017scatter}  & 89.5 & 85.4 & 87.9 & 78.8 & 75.9 & 85.4 & 74.8 & 86.1 & 100.0 & 78.1 & 90.0 & 98.6 & 85.9 \\
				JDA\cite{long2013transfer}  & 89.6 & 85.1 & 89.8 & 83.6 & 78.3 & 80.3 & 84.8 & 90.3 & 100.0 & 85.5 & 91.7 & 99.7 & 88.2\\
				JGSA\cite{zhang2017joint} & 91.4 & 86.8 & 93.6 & 84.9 & 81.0 & 88.5 & 85.0 & 90.7 & 100.0 & 86.2 & 92.0 & 99.7 & 90.0\\
				MEDA\cite{wang2018visual} & \textbf{93.4} & \textbf{95.6} & 91.1 & \textbf{87.4} & 88.1 & 88.1 & \textbf{93.2} & \textbf{99.4} & 99.4 & 87.5 & \textbf{93.2} & 97.6 & \textbf{92.8}\\
				SLPP & 91.3 & 73.6 & 86.6 & 82.6 & 72.2 & 82.8 & 71.8 & 79.5 & 100.0 & 79.2 & 88.5 & 99.3 & 84.0\\
				CAPLS(LDA) & 91.1 & 85.4 & 94.9 & 83.5 & 86.4 & 90.4 & 87.7 & 92.5 & 100.0 & 87.8 & 92.4 & 99.7 & 91.0\\ \hline
				CAPLS(Ours) & 90.8 & 85.4 & \textbf{95.5} & 86.1 & 87.1 & \textbf{94.9} & 88.2 & 92.3 & \textbf{100.0} & \textbf{88.8} & 93.0 & \textbf{100.0}& 91.8\\
				\hline
				\hline
			\end{tabular}
		\end{lrbox}
		\scalebox{0.85}{\usebox{\tablebox}}
	}
\end{table*}

\begin{table}[!htbp]
	\centering
	{
		\centering
		\caption[]{Classification Accuracy (\%) on Office31 dataset for unsupervised domain adaptation.  The feature transformation approaches use ResNet50 features and deep models use ResNet50 as backbones. Results are from the original paper except the ones labelled with $^*$ for which we report the results from the supplementary material of \cite{wang2018visual}.\\
		}
		\label{table:uda_o31}
		\begin{lrbox}{\tablebox}
			\begin{tabular}{cccccccc}
				\hline
				Method &A$\to$W & D$\to$W & W$\to$D & A$\to$D & D$\to$A & W$\to$A & Average \\ \hline
				DAN$^*$\cite{long2015learning}  & 80.5 & 97.1 & 99.6 & 78.6 & 63.6 & 62.8 & 80.4 \\
				JDDA\cite{chen2018joint} & 82.6 & 95.2 & 99.7 & 79.8 & 57.4 & 66.7 & 80.2\\
				RTN\cite{long2016unsupervised} & 84.5 & 96.8 & 99.4 & 77.5 & 66.2 & 64.8 & 81.6\\
				MADA\cite{pei2018multi} & 90.0 & 97.4 & 99.6 & 87.8 & 70.3 & 66.4 & 85.2 \\
				GTA\cite{sankaranarayanan2017generate} & 89.5& 97.9 & 99.8& 87.7 & 72.8 & 71.4& 86.5\\
				iCAN\cite{zhang2018collaborative} & 92.5 & {98.8} & \textbf{100.0} & 90.1 & 72.1 & 69.9 & 87.2 \\
				CDAN-M\cite{long2018conditional} & \textbf{93.1} & 98.6 & \textbf{100.0} & \textbf{92.9} & 71.0 & 69.3& 87.5\\
				\hline
				MEDA\cite{wang2018visual} & 86.2 & 97.2 & 99.4 & 85.3 & 72.4 & 74.0 & 85.7\\
				SLPP & 77.9 &97.4 &99.2 &80.1 & 68.4 & 66.2 & 81.5\\
				CAPLS(LDA) & 77.0 & \textbf{99.1} & 99.8 & 77.9 & 61.8 & 60.8 & 79.4 \\ \hline
				CAPLS(Ours) & 90.6 & 98.6 & 99.6 & 88.6 & \textbf{75.4} & \textbf{76.3} & \textbf{88.2}\\
				\hline
				\hline
			\end{tabular}
		\end{lrbox}
		\scalebox{0.8}{\usebox{\tablebox}}
	}
\end{table}

\begin{table*}[!htbp]
	\centering
	{
		\centering
		\caption[]{Classification Accuracy (\%) on Office-Home dataset for unsupervised domain adaptation. The feature transformation approaches use ResNet50 features and deep models use ResNet50 as backbones. Results are from the original paper except the ones labelled with $^*$ for which we report the results from the supplementary material of \cite{wang2018visual}.\\	}
		\label{table:uda_o65}
		\begin{lrbox}{\tablebox}
			\begin{tabular}{cccccccccccccc}
				\hline
				Method & A$\to$C & A$\to$P & A$\to$R & C$\to$A&C$\to$P & C$\to$R&P$\to$A & P$\to$C & P$\to$R & R$\to$A & R$\to$C & R$\to$P & Average \\ \hline
				DAN$^*$\cite{long2015learning}  & 43.6 & 57.0 & 67.9 & 45.8 & 56.5 & 60.4 & 44.0 & 43.6 & 67.7 & 63.1 & 51.5 & 74.3 & 56.3\\
				JAN$^*$\cite{long2017deep} & 45.9 & 61.2 & 68.9 & 50.4 & 59.7 & 61.0 & 45.8 & 43.4 & 70.3 & 63.9 & 52.4 & 76.8 & 58.3\\
				CDAN-M$^*$\cite{long2018conditional} & 50.6 & 65.9 & 73.4 & 55.7 & 62.7 & 64.2 & 51.8 & 49.1 & 74.5 & 68.2 & 56.9 & 80.7 & 62.8\\
				\hline
				MEDA\cite{wang2018visual} & 54.6 & 75.2 & 77.0 & 56.5 & 72.8 & 72.3 & 59.0 & 51.9 & 78.2 & 67.7 & \textbf{57.2} & 81.8 & 67.0\\
				SLPP & 49.3 & 70.5 & 74.9 & 55.7 & 68.9 & 69.7 & 57.2 & 47.3 & 75.4 & 67.5 & 53.0 & 80.5 & 64.2\\
				CAPLS(LDA) &47.1 & 72.8 & 77.6 & 57.3 & \textbf{76.5} & 78.0 & 55.8 & 47.7 & \textbf{81.7} & 65.9 & 52.6 & \textbf{84.5} & 66.5\\ \hline
				CAPLS(Ours) & \textbf{56.2} & \textbf{78.3} & \textbf{80.2} & \textbf{66.0} & 75.4 & \textbf{78.4} & \textbf{66.4} & \textbf{53.2} & 81.1 & \textbf{71.6} & 56.1 & 84.3 & \textbf{70.6}\\
				\hline
				\hline
			\end{tabular}
		\end{lrbox}
		\scalebox{0.85}{\usebox{\tablebox}}
	}
\end{table*}

In this section, we describe our experiments on three commonly used domain adaptation datasets (i.e. Office+Caltech \cite{gong2012geodesic}, Office31 \cite{saenko2010adapting} and Office-Home \cite{venkateswara2017deep}) under unsupervised and zero-shot learning conditions. The experimental results are presented and compared with those of state-of-the-art domain adaptation approaches.

\subsection{Datasets}\label{sec:dataset}
\textbf{Office+Caltech} dataset is one of the most commonly used datasets for unsupervised domain adaptation released by Gong et al. \cite{gong2012geodesic}. The dataset consists of four domains: Amazon (images downloaded from online merchants), Webcam (low-resolution images by a web camera), DSLR (high-resolution images by a digital SLR camera) and Caltech-256.  10 common classes from all four domains are used: backpack, bike, calculator, headphone, computer-keyboard, laptop-101, computer-monitor, computer-mouse, coffee-mug, and video-projector. There are 2533 images in total with 8 to 151 images per category per domain. The Decaf6 \cite{donahue2014decaf} features (activations of the 6\textit{th} fully connected layer of a convolutional neural network trained on ImageNet) are used in our experiments for a direct comparison with others.

\textbf{Office31} dataset \cite{saenko2010adapting} is also a benchmark dataset commonly used for evaluating different domain adaptation approaches. The dataset consists of three domains: Amazon, Webcam and DSLR. There are 31 common classes for all three domains containing 4,110 images in total. ResNet50 \cite{he2016deep} has been commonly used to extract features or as the backbone of deep models in literature, hence we use ResNet50 features in our experiments for this dataset.

\textbf{Office-Home} dataset \cite{venkateswara2017deep} is another dataset recently released for evaluation of domain adaptation algorithms. It consists of four different domains: Artistic images, Clipart, Product images and Real-World images. There are 65 object classes in each domain with a total number of 15,588 images. Again, we extracted ResNet50  features in our experiments for fair comparisons with others.

\subsection{Experiments on Unsupervised Domain Adaptation}\label{sec:exp_uda}
To evaluate the effectiveness of our proposed approach on unsupervised domain adaptation, we conduct comparative experiments on all three datasets. Following the standard protocols \cite{gong2012geodesic, wang2018visual}, we exhaustively select two different domains from one dataset as the source domain and target domain respectively, which allow us to have 12, 6 and 12 combinations for Office+Caltech, Office31 and Office-Home datasets respectively. We compare the performance of our approaches with those of typical state-of-the-art methods including both feature transformation and deep feature learning approaches. We follow \cite{wang2018visual} and use per-image accuracy as the evaluation metric in all our experiments.

In addition, we also report the performance of some baseline methods to evaluate how different components of our approach contribute to the final performance. To evaluate the contribution of the CAPLS strategy, we remove it from our approach and report the performance of the baseline method dubbed \textbf{SLPP}. To compare the performance of different subspace learning algorithms, we replace the SLPP with LDA and report the performance of this baseline which is named as \textbf{CAPLS(LDA)}.

\begin{figure}
	\centering
	{\includegraphics[width = 0.48\textwidth]{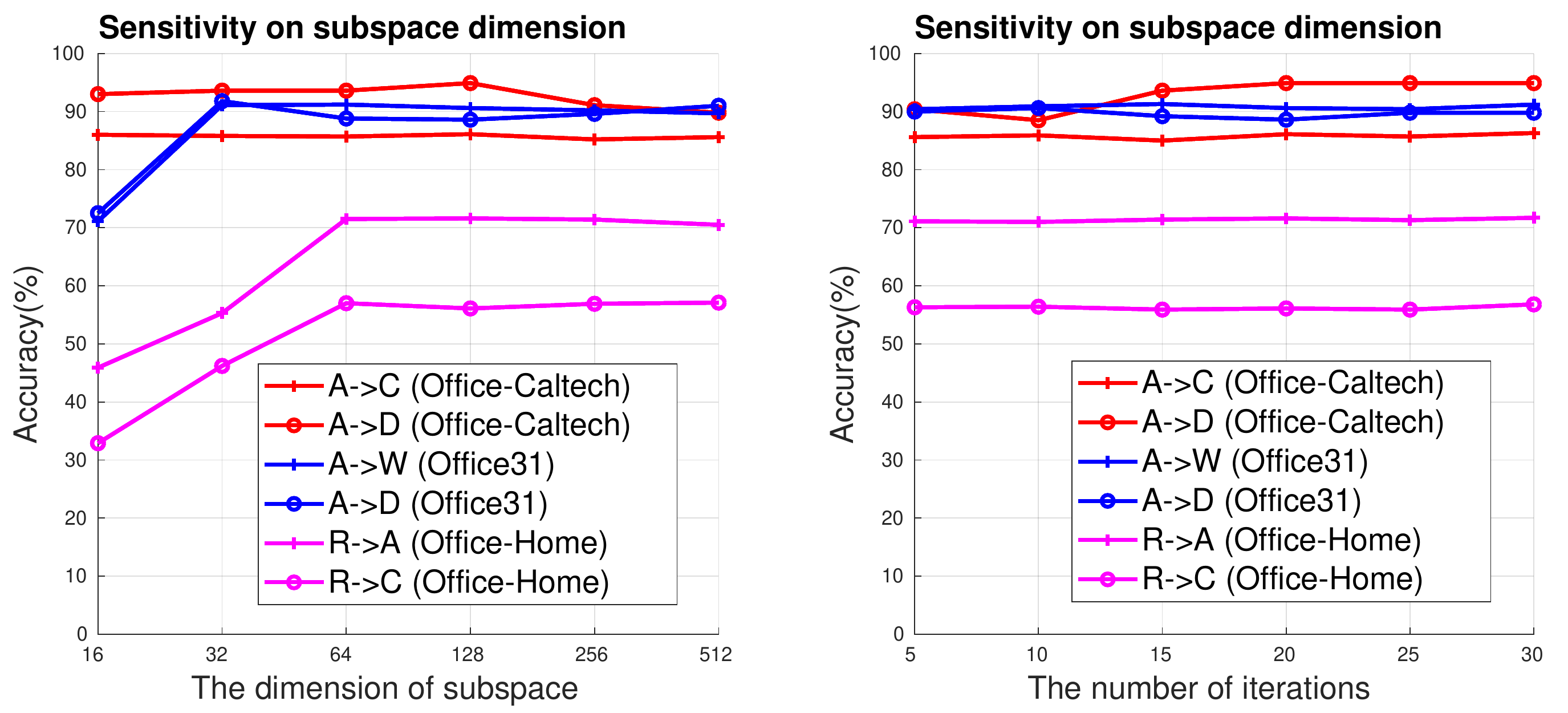}}
	{\caption{The effect of different values of $d$ (left) and $T$(right).}
		\label{fig:sensitivity}}
\end{figure}

\begin{table*}[!htbp]
	\centering
	{
		\centering
		\caption[]{Classification Accuracy (\%) on Office-Home dataset for zero-shot learning condition. We report the mean accuracy over 5 random known/unseen class splits and standard errors of the mean. \\	}
		\label{table:zsl_o65}
		\begin{lrbox}{\tablebox}
			\begin{tabular}{c|ccc|ccc|ccc}
				\hline
				\multirow{2}{*}{Method} & \multicolumn{3}{c|}{R$\rightarrow$A}& \multicolumn{3}{c|}{R$\rightarrow$C}& \multicolumn{3}{c}{R$\rightarrow$P}\\ \cline{2-10}
				& Known & Unseen & H & Known & Unseen & H &Known & Unseen & H\\
				\hline
				1NN & $64.7\pm 0.7$ &$52.3\pm 0.7$ &$57.8\pm 0.2$ & $71.6\pm 0.4$ &$32.5\pm 0.7$ &$44.7\pm 0.7$ &$87.0 \pm 0.6$ &$67.5\pm 1.0$ &$76.0\pm 0.5$ \\
				SVM & $72.1\pm 0.3$ &$58.4\pm 0.4$ &$64.5\pm 0.3$ &$66.6\pm 0.3$ &$37.7\pm 1.2$ &$48.1\pm 1.0$ &$88.4\pm 0.3$&$71.8\pm 1.5$ &$79.2\pm 0.8$\\
				LDA &$68.9\pm 0.4$ & $60.7\pm 0.7$ &$64.5\pm 0.4$ &$71.5\pm 0.6$&$42.6\pm 0.7$ &$53.3\pm 0.5$ &$88.9\pm 0.3$ &$78.0\pm 0.4$&$83.1\pm 0.2$ \\
				MR \cite{xu2017transductive} &$68.2\pm 0.9$ &$2.1\pm 0.3$ &$4.0\pm 0.6$ &$71.9\pm 1.0$ &$2.3\pm 0.4$ &$4.5\pm 0.8$ &$88.1\pm 0.4$ &$3.7\pm 0.7$ &$7.0\pm 1.3$ \\
				BiDiLEL\cite{wang2017zero} &$\mathbf{73.9\pm 1.1}$ & $9.0\pm 1.2$&$15.9\pm 1.8$ &$\mathbf{74.3\pm 0.8}$ &$8.3\pm 0.6$ &$14.9\pm 1.0$ &$\mathbf{89.4\pm 0.3}$ &$14.0\pm 1.4$&$24.2\pm 2.0$ \\ \hline
				Ours &$72.2\pm 0.7$ &$\mathbf{64.6\pm 0.9}$ &$\mathbf{68.2\pm 0.5}$ &$73.8\pm 0.9$ &$\mathbf{46.5\pm 1.0}$ &$\mathbf{57.0\pm 0.7}$ &$89.3\pm 0.2$ &$\mathbf{78.6\pm 0.5}$ &$\mathbf{83.6\pm 0.2}$\\
				\hline
				\hline
			\end{tabular}
		\end{lrbox}
		\scalebox{0.9}{\usebox{\tablebox}}
	}
\end{table*}

Our approach consists of two hyper-parameters: the dimensionality of subspace $d$ and the number of iterations $T$. We conduct an experiment to investigate how our approach is sensitive to these two hyper-parameters. Firstly, we fix $T=20$ and set $d=\{16, 32, 64, 128, 256, 512\}$ respectively and get results presented in Figure \ref{fig:sensitivity} (left). It is obvious that when $d$ is greater than 64, we can achieve stable accuracy for all three datasets.  Secondly, we set $d=128$ and $T=\{5,10,15,20,25,30\}$ respectively and get results shown in Figure \ref{fig:sensitivity} (right), which indicates our approach is not sensitive to the number of iterations $T$. As a result, we set fixed values as $d=128$ and $T=20$ across all the experiments in this section.

Table \ref{table:uda_o10} shows the results of comparative experiments on Office+Caltech dataset under the unsupervised domain adaptation condition. For a fair comparison, the results of all methods are based on Decaf6 features. By comparing our approach with two baseline methods, i.e., SLPP and CAPLS(LDA), we can see that both the SLPP based joint subspace learning and the confidence-aware pseudo label selection scheme play important roles in achieving good performance for unsupervised domain adaptation. We compare our approach with three deep learning based methods (i.e. DCORAL\cite{sun2016deep}, DDC\cite{tzeng2014deep} and DAN\cite{long2015learning}, GTA\cite{sankaranarayanan2017generate}). Our approach achieves superior accuracy than the deep feature learning models in most tasks and a superior average accuracy of 91.8\%. When compared with other feature transformation approaches, our approach ranks the \textbf{second} in terms of the average accuracy over 12 tasks with slightly worse performance than MEDA \cite{wang2018visual} which is a combination of manifold feature learning and dynamic distribution alignment techniques.

Table \ref{table:uda_o31} shows the results of comparative experiments on Office31 dataset under the unsupervised domain adaptation condition. For a fair comparison, all the feature transformation approaches use ResNet50 features and all the deep learning methods use ResNet50 as their backbones. Our proposed approach achieves the best performance of 88.2\% in terms of the average accuracy over 6 tasks, outperforming seven state-of-the-art deep feature learning methods (i.e., DAN\cite{sun2016deep}, JDDA\cite{chen2018joint}, RTN\cite{long2016unsupervised}, MADA\cite{pei2018multi}, iCAN\cite{zhang2018collaborative} and CDAN-M\cite{long2018conditional}) and the competitive feature transformation approach MEDA \cite{wang2018visual}. The comparison with two baseline methods also indicate the effectiveness of SLPP as the subspace learning algorithm and the necessity of CAPLS for unsupervised domain adaptation.

Table \ref{table:uda_o65} displays the results of comparative experiments on Office-Home dataset. Since this is a relatively new dataset, there are very few results reported on it. We compare with the results from supplementary materials of \cite{wang2018visual}. Again ResNet50 is used for feature extraction or backbone networks for a fair comparison. We can see that our proposed approach outperforms others significantly with the average accuracy of 70.6\%. In addition, the two baseline methods have worse results than our full model which validates the effectiveness of our framework.

Although our method achieves state-of-the-art results when deep features are employed, its performance degrades when hand-crafted features are used on Office-10 dataset for which the experimental results are not presented in this paper. One possible reason is our method favours the feature space with smaller domain shift which provides more correct pseudo-labels at the beginning of training.
\subsection{Experiments on Zero-Shot Learning}\label{sec:exp_zsl}

To evaluate the proposed framework under a zero-shot learning condition, we conduct experiments on Office-Home dataset which consists of a sufficient number (65) of classes for ZSL. The source and target domain features are extracted by ResNet50 pre-trained on ImageNet. To simulate a ZSL scenario, we randomly select 35 classes as known classes for which there are labelled target data during subspace learning and the rest 30 classes are ``unseen" for which no target data are available for learning. In our experiments, the ``RealWorld" domain serves as source domain whilst ``Art", ``Clipart" and ``Product" domains serve as target domain respectively. The reason is ``RealWorld" domain data are usually easier to collect than other three domains in realistic applications. As a result, we have three tasks: R$\rightarrow$A, R$\rightarrow$C and R$\rightarrow$P. In each task, all source domain (``RealWorld") data are used for training. In addition, half of the target data in each class are reserved for testing and the other half for training only if they belong to known classes.

We use fixed training/test data split for target domain data and randomly generate 5 known/unseen class splits for a thorough evaluation \footnote{https://github.com/hellowangqian/domain-adaptation-capls.}. Considering the unbalanced number of test images in different classes, we use per-class mean accuracy for evaluation in this experiment. Following the common way in evaluating generalized zero-shot learning algorithms \cite{xian2018zero}, we report the mean accuracy on known classes $Acc^{known}$ and unseen classes $Acc^{unseen}$ respectively as well as their harmonic mean $H=2*Acc^{known}*Acc^{unseen}/(Acc^{known}+Acc^{unseen})$.

We compare our approach with two baseline methods: 1 Nearest Neighbour (\textbf{1NN}) and Support Vector Machine (\textbf{SVM}). All source domain data and labelled target domain data are combined for training without considering any domain alignment in these two baseline methods. We also compare with state-of-the-art zero-shot learning algorithms. As mentioned in Section \ref{sec:prob}, traditional zero-shot learning algorithms can only handle class-level representations in the source domain. To adapt them to our problem, we calculate class-level source domain representations by averaging the source domain samples belonging to the same class. After this adaptation, we apply two zero-shot learning algorithms, i.e., bidirectional latent embedding learning (\textbf{BiDiLEL}) \cite{wang2017zero} and manifold regularized ridge regression (\textbf{MR}) \cite{xu2017transductive} to our problem for the comparison. We also replace the SLPP with \textbf{LDA} in our framework to validate the importance of preserving data structure when learning subspace of favourable properties.

Table \ref{table:zsl_o65} shows the results of comparative experiments on Office-Home dataset under zero-shot learning condition. We can see that the adapted zero-shot learning algorithms fail in this problem with very low classification accuracy in terms of unseen classes though BiDiLEL achieves the best accuracy on known classes. As a result, two adapted zero-shot learning algorithms perform much worse than the baseline methods in terms of the harmonic mean $H$. Our proposed framework works reasonably well on all three tasks with high classification accuracy for both known and unseen classes. With the use of SLPP, our approach achieves the best classification accuracy on unseen classes and second best on known classes. The experimental results provide evidence that our proposed framework is a promising solution to domain adaptation problems under zero-shot learning conditions.
\section{Conclusion}
In this paper, we propose a unified framework for visual domain adaptation under unsupervised and zero-shot learning conditions. A domain-invariant and discriminative subspace is learned by SLPP using labelled/pseudo-labelled data from both domains. A confidence-aware pseudo label selection scheme is proved to be effective for choosing proper pseudo-labelled target samples in the iterative learning. Our experimental results on unsupervised domain adaptation and zero-shot learning problem prove that the proposed approach achieves state-of-the-art performance.
\bibliographystyle{IEEEtran}
\bibliography{ijcnn2019}

\end{document}